\documentclass[letterpaper]{article}
\usepackage{flairs}%aaai
\usepackage{times}
\usepackage{helvet}
\usepackage{courier}
\usepackage{amsmath,amssymb,amsfonts}
\usepackage{algorithmic}
\usepackage{graphicx}
\usepackage{textcomp}
\usepackage{xcolor}
\usepackage{url}
\usepackage{xurl}
\begin{document}
% This file is an adoption of the style file for AAAI Press 
% proceedings, working notes, and technical reports.  This file is made 
% with minimal changes by explicit permission from AAAI.
\title{TemporalAugmenter: An Ensemble Recurrent Based Deep Learning Approach for Signal Classification}
%\author{Anonymous for Reviewing\\
%	Anonymous for reviewing\\
%	Anonymous for reviewing\\
%	Anonymous for reviewing\\
	%\And Author 3\\
	%Affiliation 3\\
	%Address Line 3.1\\
	%Address Line 3.2\\
%}
\author{Nelly Elsayed\\
	School of IT\\
	University of Cincinnati\\
	elsayeny@ucmail.uc.edu\\
	\And Constantinos L. Zekios\\
Dep. of Elect. \& Comp. Eng.\\
	Florida International University\\
kzekios@fiu.edu\\
		\And Navid Asadizanjani\\
	Dep. of Elect. \& Comp. Eng.\\
	University of Florida\\
	nasadi@ece.ufl.edu\\
		\And Zag ElSayed\\
	School of IT\\
	University of Cincinnati\\
elsayezs@ucmail.uc.edu\\
}
\maketitle
\begin{abstract}
\begin{quote}
Ensemble modeling has been widely used to solve complex problems as it helps to improve overall performance and generalization. In this paper, we propose a novel TemporalAugmenter approach based on ensemble modeling for augmenting the temporal information capturing for long-term and short-term dependencies in data integration of two variations of recurrent neural networks in two learning streams to obtain the maximum possible temporal extraction. Thus, the proposed model augments the extraction of temporal dependencies. In addition, the proposed approach reduces the preprocessing and prior stages of feature extraction, which reduces the required energy to process the models built upon the proposed TemporalAugmenter approach, contributing towards green AI. Moreover, the proposed model can be simply integrated into various domains including industrial, medical, and human-computer interaction applications. Our proposed approach empirically evaluated the speech emotion recognition, electrocardiogram signal, and signal quality examination tasks as three different signals with varying complexity and different temporal dependency features.
\end{quote}
\end{abstract}

\noindent  Ensemble modeling is one of the solutions to overcome the model overfitting and improve the model performance by integrating multiple individual learning steams to create a robust and accurate predictive model, especially for complex tasks~\cite{ganaie2022ensemble,sagi2018ensemble}. The ensemble modeling concept improves the overall model generalization, robustness, and stability and improves the overall model accuracy~\cite{arpit2022ensemble,ortega2022diversity,zhang2019enhancing}. Ensemble modeling has been applied in various applications including time series classification, speech recognition, image classification~\cite{karim2019insights,kourentzes2014neural,elsayed2018deep}, natural language processing~\cite{sangamnerkar2020ensemble,liu2019ensembles,jia2023review}, events detection and recognition in videos~\cite{adewopo2023smart,yu2020deep,xu2018ensemble,nanni2014ensemble}, anomaly detection~\cite{han2021gan,zhao2015ensemble,vanerio2017ensemble}, security of IoT devices~\cite{tsogbaatar2021iot,elsayed2023iot,alotaibi2023ensemble}, and medical applications~\cite{west2005ensemble,priyadharshini2023hybrid,liu2020novel}. 

\begin{figure*}[htbp]
	\centerline{\includegraphics[width=13cm, height= 3 cm]{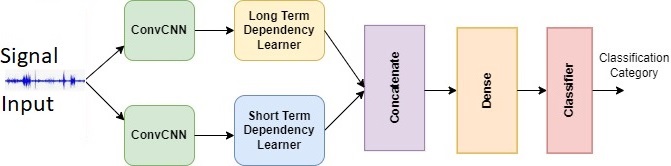}}
	\caption{The proposed ensemble TemporalAugmenter approach for both long-term and short-term dependencies capturing in temporal data.}
	\label{model_diagram}
\end{figure*}

Several ensemble modeling techniques in deep learning include boosting, bagging, stacking, negative correlation-based deep ensemble method, explicit and implicit ensembles, homogeneous and heterogeneous ensembles, and decision fusion strategies~\cite{ganaie2022ensemble}. The boosting technique is based on training the models sequentially, where each subsequent model focuses on solving the weaknesses of the previous model. Then, the highest weights are assigned to instances that were misclassified in the previous stages. That enables the model to learn from its mistakes, which helps to gradually improve the overall accuracy~\cite{drucker1994boosting,ferreira2012boosting,mosavi2021ensemble}. The bagging technique is based on training multiple models on subsets of the training data in parallel. Then, the final prediction is performed via aggregating all the models' predictions by taking a vote or the average~\cite{altman2017ensemble,galar2011review}. The stacking technique is based on combining multiple base model predictions by using an additional model called a meta-learner that is responsible for learning how to perform the best prediction based on the predictions of the multiple base models~\cite{brownlee2021ensemble}. The negative correlation-based deep ensemble technique is based on training models that are negatively correlated by training the models in a way that aims to make different predictions, leading to promoting diverse predictions and reducing redundancy~\cite{ganaie2022ensemble}. The explicit technique combines multiple distinct models and performs the training explicitly~\cite{ganaie2022ensemble}. The implicit technique uses model uncertainty estimation within a single model or a dropout to create an implicit ensemble effect on the entire model~\cite{seijo2017ensemble}. The homogeneous technique combines multiple models of the same type to enhance a prediction concept. The heterogeneous technique consists of different models in the concept to enhance the diverse learning strategy~\cite{seijo2017ensemble}. The decision fusion technique is based on using multiple models to combine their final prediction decisions based on simple or complex methodology such as averaging, voting, or weights assigned to individual models's predictions ~\cite{ponti2011combining,ganaie2022ensemble,hassan2007decisions}.

There are different types of data depending on the primary source of data capturing, including discrete and sequential datasets~\cite{dietterich2002machine,chmielewski1996global}. The sequential data can be categorized into temporal data, where data points are collected, observed, and recorded at the same specific time intervals (e.g., videos, voice recording, time series, biological signal), and sequential (non-temporal) data that involves sequences where the order is significant. However, the observation time is not considered (e.g., text data, ordered events, and DNA sequences). The temporal data is complex due to the temporal information and the point-in-time information that the learning model must capture to perform the required task on the data. Thus, not all traditional learning models can solve temporal data based problems and tasks. Recurrent neural based architectures are the most suitable for capturing the temporal dependency information carried in the temporal data. There are several recurrent neural network based architectures such as the recurrent neural network (RNN) long short-term memory (LSTM) and its different variants~\cite{greff2017lstm,elsayed2022litelstm,gers2002learning}, the gated recurrent unit (GRU) and different variants~\cite{chung2014empirical,dey2017gate}, the LiteLSTM~\cite{elsayed2023litelstm}, and the minimal gated unit MGU~\cite{zhou2016minimal}. Each recurrent based network has its strengthes in capturing the long-term or short-term temporal dependencies. However, with the complex data, they require additional preprocessing or support for feature extracting to enhance the overall performance of capturing the point-to-time information. 

Thus, in this paper, we propose a novel ensemble approach for augmenting the temporal information in temporal data based long-term and short-term dependencies capturing architectures: long short-term memory (LSTM) and the gated recurrent unit (GRU) in two streams that are capable of improving the overall performance. Moreover, we performed an empirical investigation of the influence of the convolutional neural network as a feature extractor prior to short-term temporal dependencies capturing architecture on improving the model performance; in addition, it eliminates the requirement of the data preprocessing stage. Finally, we employed the proposed approach on three different temporal tasks from different data sources and varied complexity, including speech emotion recognition, electrocardiogram classification, and radar signal quality classification, to analyze and validate the proposed approach concepts.

\begin{figure}[t]
	\centerline{\includegraphics[width=3cm, height= 10 cm]{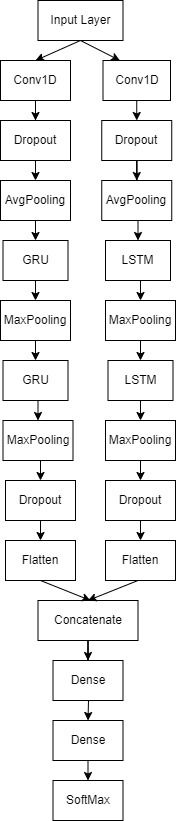}}
	\caption{The proposed model based on the TemporalAugmenter appoach implementation layers and their orders.}
	\label{model_shape}
\end{figure}

\section{Temporal Augmenter Ensemble Approach}

The proposed temporal augmenter approach is shown in Fig.~\ref{model_diagram}~\cite{elsayed2022litelstm}. The proposed approach consists of two stacked streams. The first stream employs a recurrent neural network architecture that is capable of extracting and learning the long-term temporal dependencies in the data. The second stream employs the recurrent neural network architecture that is capable of extracting and learning the short-term dependencies in the data. Adding one convolutional layer before each recurrent stream would improve the spatial features from the data, contributing to improving the overall approach performance at both temporal and spatial dependencies extraction and learning. Integrating the convolutional layer recurrent architectures has shown empirically and overall higher performance in several 1D applications including~\cite{elsayed2022speech,pan2020water,sajjad2020novel,elsayed2023iot}. In this proposed approach, we empirically found that the optimal integration between the temporal dependencies extraction in the recurrent model and the convolution neural network (CNN) can be found while using the CNN as only one layer for extracting features prior to the recurrent network. Thus, the proposed approach reduces the required computations for the preprocessing of the signal due to the capability of the one-layer CNN to extract sufficient features, eliminating the requirements of signal preprocessing.

For the long-term dependency learning stream, in this approach, we employ the long short-term memory (LSTM) architecture as the main component for long-term dependency learning~\cite{greff2016lstm}. The memory cell of the LSTM provided the capability of memorizing the long-term dependencies due to maintaining the long-term dependencies information in the learning stream. By reflecting the forget gate in the memory cell, the memory maintains the time dependencies that have long-term effects through the time in the memory, leading the LSTM to maintain a robust memorization of long-term dependencies from the data.

For the short-term dependencies learning stream, we selected the gated recurrent unit (GRU) as the primary component to learn the short-term dependencies in the data. The GRU is a smaller recurrent neural network architecture that consists of two gates: update $z$ and reset gates $r$~\cite{greff2017lstm}. The main concept was to share the weights between two gates at the LSTM (input and forget) gates into one update gate and remove the memory cell to produce a smaller budget recurrent architecture that can be employed in applications where the sequences are small (e.g., short-term dependencies)~\cite{chung2014empirical}. In addition, the GRU eliminates the output squashing function and the constant error carrocel (CEC) compared to the LSTM~\cite{chung2014empirical,elsayed2023litelstm}. The reset gate in the GRU maps the output gate of the LSTM. Thus, the GRU requires less budget to implement compared to the LSTM, and it is capable of learning short-term dependencies efficiently. Thus, the GRU has shown significant results and outperformed the LSTM in several applications where the time dependency in the data is short-term, such as~\cite{shen2018deep,elsayed2019gated,gao2020gated,yiugit2021automatic,elsayed2021arrhythmia,golmohammadi2017gated,wang2021speech,elsayed2023iot,jakubik2018evaluation,al2021deep}.

Models that are based on our proposed TemporalAugmenter approach aim to employ both the LSTM as the long-term dependencies learning architecture with the GRU as a short-term dependencies learner into a model that is capable of capturing the long-term dependencies in temporal data as well as the short-term dependencies. Thus, the proposed TemporalAugmenter approach-based model can exceed the state-of-the-art models in multiple applications with different temporal data sources.
\begin{figure}[t]
	\centerline{\includegraphics[width=8.5cm, height=3.5 cm]{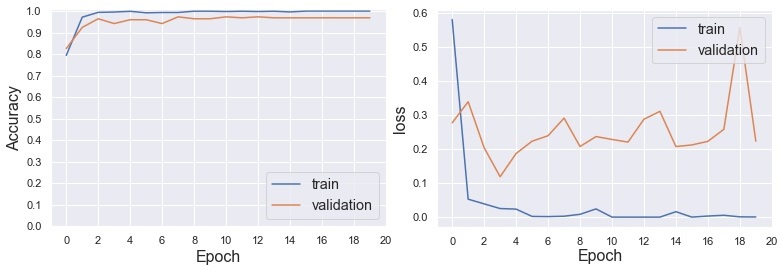}}
	\caption{The proposed TemporalAugmenter approach training versus validation accuracy (left) and loss (right) diagrams over the TESS dataset.}
	\label{tess_train_validate}
\end{figure}

\begin{table}[t]
	\caption{The proposed model overall statistics for speech emotion recognition from the TESS dataset.}
	\begin{center}
		\scriptsize
		\begin{tabular}{|l|c|}
			\hline
			\textbf{Merits}& \textbf{Value}\\
			\hline
			95\% CI      &                    (0.9783,0.9967)\\
			Accuracy &99.64\%\\
			F1 Score&                                                         0.9875\\
			
			False Negative Rate&                                                        0.0125\\
			False Positive Rate&                                                    0.00208\\
			True Negative Rate&                                                     0.99792\\
			
			True Positve Rate      &                                                     0.9875\\
			Kappa              &                                               0.9854\\
			Kappa 95\% CI         &                                             (0.97465,0.99615)\\
			Kappa Standard Error              &                                0.00548\\
			Total params& 8,027,819   \\
			Trainable params& 8,027,819 \\
			Non-trainable params& 0    \\
			\hline
		\end{tabular}
		\label{results_tess_model}
	\end{center}
\end{table}

%\begin{figure}[htbp]
%	\centerline{\includegraphics[width=5cm, height=5 cm]{confusion_tess.png}}
%	\caption{The confusion matrix of the proposed TemporalAugmenter approach over the TESS dataset.}
%	\label{confusion_Tess}
%\end{figure}

\begin{table*}[htbp]
	\caption{The proposed model statistics over the seven speech emotion categories of the TESS dataset.}
	\begin{center}
		\scriptsize
		\begin{tabular}{|l|ccccccc|}
			\hline
			\textbf{Statistical}&\multicolumn{7}{|c|}{\textbf{Emotion Category}} \\
			\cline{2-8} 
			\textbf{Analysis} & \textbf{\textit{Angry}}& \textbf{\textit{Disgust}}& \textbf{\textit{Fear}}
			& \textbf{\textit{Happiness}} & \textbf{\textit{Surprise}}& \textbf{\textit{Sadness}}& \textbf{\textit{Neutral}} \\
			\hline
			Accuracy & 100\%       &    99.643\%      &  99.464\%     &   99.821\%     &   99.643\%    &    99.286\%    &    99.643\%     \\
			F1 Score &  1.0     &       0.9878     &    0.97902     &   0.99355    &    0.9875     &    0.97826   &     0.98571 \\
			AUC &   1.0    &        0.99286     &   0.97945   &     0.99359  &      0.99792   &     0.99131   &     0.99184   \\
			%	(Area under the ROC curve)  
			Error rate & 0.0   &        0.00357     &  0.00536      & 0.00179     &  0.00357     &  0.00714   &    0.00357 \\
			False Negative Rate&  0.0       &    0.0122    &    0.0411       & 0.01282  &     0.0   &        0.01099 &      0.01429 \\
			False Positive Rate& 0.0  &   0.00209   &    0.0  &  0.0  &0.00416  &   0.0064  &   0.00204 \\
			Specificity&    1.0  &    0.99791  & 1.0   & 1.0 &  0.99584   & 0.9936   & 0.99796    \\
			Sensitivity & 1.0      &     0.9878    &    0.9589    &    0.98718    &   1.0      &     0.98901     &  0.98571    \\
			%	OP (optimized precision)& 1.0         &  0.99134     & 0.97366     &  0.99176    &   0.99435   &    0.99054    &   0.99026\\
			\hline
		\end{tabular}
		\label{class_level_tess}
		
	\end{center}
\end{table*}

\begin{table}[t]
	\caption{Comparison between the proposed model and the state-of-the-art models for speech emotion recognition over the TESS dataset.}
	\begin{center}
		\scriptsize
		\begin{tabular}{|p{2.9cm}|p{3.2cm}|p{0.9cm}|}
			\hline
			%		\textbf{Model}&\multicolumn{7}{|c|}{\textbf{Emotion Category}} \\
			%		\cline{2-8} 
			\textbf{Model} & \textbf{Method} & \textbf{Acc.}\\
			\hline
			\cite{venkataramanan2019emotion}& Combining 2D CNN and Global Avg. Pooling &66.00\%\\
			\hline
			\cite{sundarprasad2018speech} & Combining PCA, SVM, Mel-Frequeny, and Cepstrum Features & 90.00\%\\
			\hline
			\cite{krishnan2021emotion} & SoA Classsifier and Entropy Features from Principle IMF modes & 93.30\%\\
			\hline
			\cite{lotfidereshgi2017biologically}& Liquid State Machine &82.35\%\\
			\hline
			\cite{zhang2013speech} & Kernel Isomap & 80.85\%\\
			\hline
			\cite{zhang2013speech} & PCA& 72.35\%\\
			\hline
			\cite{bhargava2013improving} &Artificial Neural Nets& 80.600\%\\
			\hline
			\cite{bhargava2013improving} &SVM& 80.270\%\\
			\hline
			\cite{elsayed2022speech}& 1DCNN and GRU & 94.285\%\\
			\hline
			\cite{parry19_interspeech}& CNN and LSTM& 49.48\%\\
			\hline
			\cite{zhao2019speech}& 2D-CNN and LSTM& 70.00\%\\
			\hline
			\textbf{Our} & \textbf{TemporalAugmenter} & \textbf{98.75\%}\\
			\hline
		\end{tabular}
		\label{tess_compare}
	\end{center}
\end{table}

\begin{figure}[t]
	\centerline{\includegraphics[width=8.5cm, height=3.5 cm]{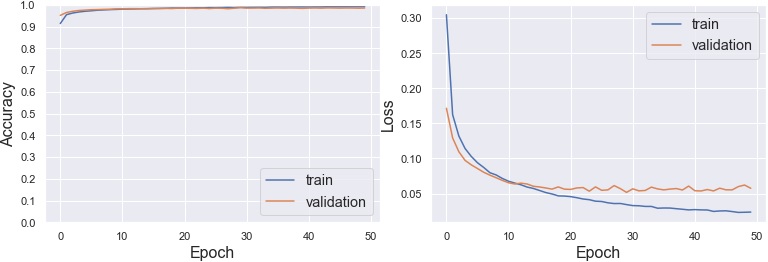}}
	\caption{The proposed TemporalAugmenter approach training versus validation accuracy (left) and loss (right) diagrams over the MIT-BIH dataset.}
	\label{mih_train_val}
\end{figure}

\begin{table}[t]
	\caption{The proposed model overall statistics for MIT-BIH dataset.}
	\begin{center}
		\scriptsize
		\begin{tabular}{|l|c|}
			\hline
			\textbf{Merits}& \textbf{Value}\\
			\hline
			95\% CI      &                    (0.98293,0.98619)\\
			Accuracy &98.456\%\\
			F1 Score&                                                       0.98456\\
			
			False Negative Rate&                                                       0.01544\\
			False Positive Rate&                                                    0.00208\\
			True Negative Rate&                                                     0.99614\\
			
			True Positve Rate      &                                                   0.00386\\
			Kappa              &                                              0.98456\\
			Kappa 95\% CI         &                                      (0.94317,0.95404)\\
			Kappa Standard Error              &                                   0.00277\\
			Total params& 52,073 \\
			Trainable params&52,073\\
			Non-trainable params& 0    \\
			\hline
		\end{tabular}
		\label{results_mit_model}
	\end{center}
\end{table}
\begin{table*}[htbp]
	\caption{The proposed model statistics over the five categories of the MIT-BIH dataset.}
	\begin{center}
		\scriptsize
		\begin{tabular}{|l|ccccc|}
			\hline
			\textbf{Statistical}&\multicolumn{5}{|c|}{\textbf{ECG Category}} \\
			\cline{2-6} 
			\textbf{Analysis} & \textbf{\textit{N}}& \textbf{\textit{S}}& \textbf{\textit{V}}
			& \textbf{\textit{F}} & \textbf{\textit{Q}} \\
			\hline
			Accuracy &    98.726\%     &   99.287\%   &     99.415\%   &     99.694\%    &    99.79\%   \\
			F1 Score &        0.99232  &    0.84942 &    0.95586  &   0.77888  &   0.9856   \\
			AUC &     0.97164   &   0.89475&    0.97698 &  0.86367 &   0.98913          \\
			%	(Area under the ROC curve)  
			Error rate &    0.01274   &    0.00713     &  0.00585  &    0.00306   & 0.0021     \\
			False Negative Rate&        0.00453    &    0.20863   &     0.04282    &    0.2716     &    0.02114     \\
			False Positive Rate&     0.0522    &   0.00187  &   0.00323   &  0.00106   &  0.00059  \\
			Specificity&    0.9478   &    0.99813   &   0.99677  &   0.99894   &  0.99941   \\
			Sensitivity & 0.99547  &   0.79137  &   0.95718  & 0.7284&  0.97886   \\
			%	OP (optimized precision)& 1.0         &  0.99134     & 0.97366     &  0.99176    &   0.99435   &    0.99054    &   0.99026\\
			\hline
		\end{tabular}
		\label{class_level_mit}
		
	\end{center}
\end{table*}

\begin{table}[t]
	\caption{Comparison between the proposed model and the state-of-the-art models for MIT-BIH dataset.}
	\begin{center}
		\scriptsize
		\begin{tabular}{|p{2.9cm}|p{3.2cm}|p{0.90cm}|}
			\hline
			%		\textbf{Model}&\multicolumn{7}{|c|}{\textbf{Emotion Category}} \\
			%		\cline{2-8} 
			\textbf{Model} & \textbf{Method} & \textbf{Acc.}\\
			\hline
			\cite{martis2013application}&	DWT and SVM&93.8\%	\\
			\hline
			\cite{elsayed2020simple} &	DWT and Random Forest &94.6\%	\\
			\hline
			\cite{asl2008support} &	DWT and LDA and RR&94.2\%	\\
			\hline
			\cite{osowski2001ecg}&	Hybrid fuzzy NN& 96.1\%	\\
			\hline
			\cite{acharya2017deep}& CNN and Augmentation& 93.5\%		\\
			\hline
			
			\cite{kachuee2018ecg}&	Deep residual CNN  &93.4\%		\\
			\hline
			\cite{elsayed2020simple}& ELM& 96.4\%	\\
			\hline
			\cite{martis2013cardiac}&SVM with RBF Kernel & 93.5\%	\\
			\hline
			\cite{zhou2017premature}& CNN and LSTM	&98.03\%	\\
			\hline
			\cite{acharya2017deep}&Daubechies Wavelet&94.3\%\\
			
			\hline
			\textbf{Our} & \textbf{TemporalAugmenter} & \textbf{98.45\%}\\
			\hline
		\end{tabular}
		\label{mit_compare}
	\end{center}
\end{table}

\begin{figure}[t]
	\centerline{\includegraphics[width=8.5cm, height=3.5 cm]{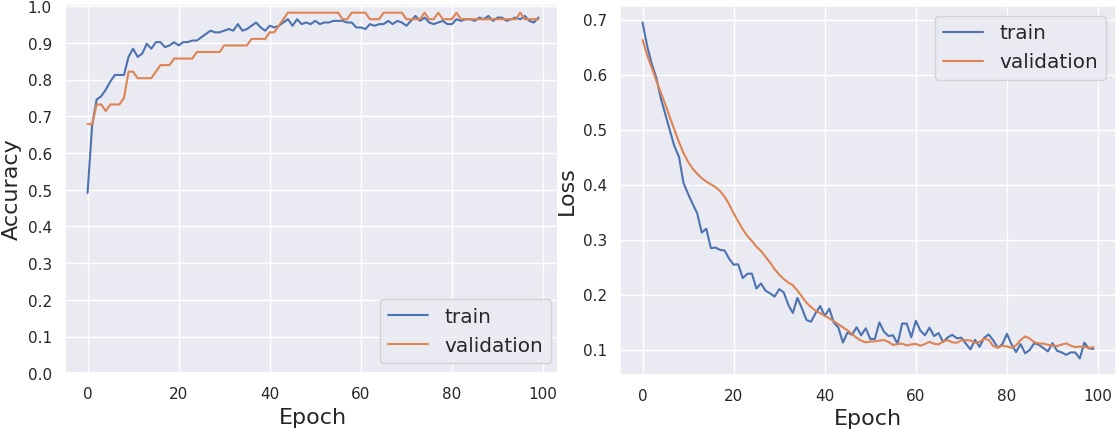}}
	\caption{The proposed TemporalAugmenter approach training versus validation accuracy (left) and loss (right) diagrams over the Radar Ionosphere Depletion dataset.}
	\label{radar_train_validate}
\end{figure}
\begin{table}[t]
	\caption{The proposed model overall statistics for the Radar Ionosphere Depletion dataset.}
	\begin{center}
		\scriptsize
		\begin{tabular}{|l|c|}
			\hline
			\textbf{Merits}& \textbf{Value}\\
			\hline
			95\% CI      &                    (0.91095, 1.0)\\
			Accuracy &95.775\%\\
			F1 Score&                                                      0.95775\\
			
			False Negative Rate&                                                      0.04225\\
			False Positive Rate&                                                 0.04225\\
			True Negative Rate&                                                    0.95775\\
			
			True Positve Rate      &                                                 0.95775\\
			Kappa              &                                              0.90839\\
			Kappa 95\% CI         &                                      (0.80693,1.00)\\
			Kappa Standard Error              &                                   0.05176\\
			Total params& 21,214\\
			Trainable params&21,214\\
			Non-trainable params& 0    \\
			\hline
		\end{tabular}
		\label{results_radar_model}
	\end{center}
\end{table}

\begin{table}[htbp]
	\caption{The proposed model statistics over the two categories of the Radar Ionosphere Depletion dataset.}
	\begin{center}
		\scriptsize
		\begin{tabular}{|l|cc|}
			\hline
			\textbf{Statistical}&\multicolumn{2}{|c|}{\textbf{Radar Signal Category}} \\
			\cline{2-3} 
			\textbf{Analysis} & \textbf{\textit{Class 0}}& \textbf{\textit{Class 1}}\\
			\hline
			Accuracy &   0.95775    &   0.95775   \\
			F1 Score &    0.96703   &    0.94118      \\
			AUC &       0.94444    &   0.94444      \\
			%	(Area under the ROC curve)  
			Error rate &0.04225    &   0.04225     \\
			False Negative Rate&       0.0      &     0.11111         \\
			False Positive Rate&     0.11111    &   0.0\\
			Specificity&    0.88889     &  1.0  \\
			Sensitivity & 1.0     &      0.88889   \\
			%	OP (optimized precision)& 1.0         &  0.99134     & 0.97366     &  0.99176    &   0.99435   &    0.99054    &   0.99026\\
			\hline
		\end{tabular}
		\label{class_level_radar}
		
	\end{center}
\end{table}

\begin{table}[t]
	\caption{Comparison between the proposed model and the state-of-the-art models for Radar Ionosphere Depletion dataset.}
	\begin{center}
		\scriptsize
		\begin{tabular}{|p{2.9cm}|p{3.2cm}|p{0.90cm}|}
			\hline
			%		\textbf{Model}&\multicolumn{7}{|c|}{\textbf{Emotion Category}} \\
			%		\cline{2-8} 
			\textbf{Model} & \textbf{Method} & \textbf{Acc.}\\
			\hline
			\cite{basheer2024autonomous}	&	AADS& 76.13\%\\
			\hline
			\cite{basheer2024autonomous}	& Streaming TEDA& 52.38\%\\
			\hline
			\cite{basheer2024autonomous}	&MAD & 22.35\%\\
			\hline
			\cite{basheer2024autonomous}		&	xStream&21.38\%\\
			\hline
			\cite{basheer2024autonomous}	&	RRCF & 12.65\%\\
			\hline
			\cite{sigillito1989classification}  & linear perceptron& 90.67\%\\
			\cite{sigillito1989classification} 	&	MLFN& 83.8\%\\
			\hline
			\textbf{Our} & \textbf{TemporalAugmenter} & \textbf{95.775\%}\\
			\hline
		\end{tabular}
		\label{radar_compare}
	\end{center}
\end{table}

\section{Experiments Setup}
In our experiment, we targeted three different tasks: speech emotion recognition, electrocardiogram (ECG or EKG) classification, and radar signal quality classification tasks. These tasks are based on three different source of temporal data that vary in complexity and features behavior through time. 

In our experiments, we implemented the models based on the proposed TemporalAugmenter approach. Figure~\ref{model_shape} shows the model layers and order, which has been implemented based on the proposed TemporalAugmenter ensemble approach. The major differences between the experiments are in the number of epochs, batch size, optimization function, input size, and the number of classification categories. For the implementation, we used a computer with an Intel(R) Core(TM) i-9 CPU @ 3.00 GHz processor with 32-GB memory and NVIDIA GeForce RTX 2080 Ti graphics card. For the implementation, we used Tensorflow 2.4.0, Numpy 1.19.5, Pandas 1.2.4, Librosa 0.9.1, and Python 3.3.8 on a Windows 10 OS computer.

\section{Task I: Speech Emotion Recognition}

\subsection{Dataset Description}
In this task, we used audio-based speech emotion datasets, and we implemented the models to use the audio data directly without any mapping to spectrograms or assigning images or videos with the data. Thus, the proposed TemporalAugmenter approach aims to capture the temporal and spatial features from the spoken speech to determine the individual's emotions during the speech.

We used the Toronto Emotional Speech Set (TESS) dataset benchmark~\cite{dupuis2010toronto}. This data was recorded in the Toronto area by two actresses who have English as their first spoken language. This dataset consists of 2800 stimuli that represent seven different emotion categories. These emotions are anger, disgust, fear, happiness, surprise/pleasant, sadness, and neutral. The major advantage of this dataset is that the dataset is balanced between the number of stimuli among each of the seven classes. 

\subsection{Results and Analysis}
In this experiment, the data was split to train, validate, and test with a ratio of 70\%:10\%:20\%, respectively. The data has been scaled using standardscaler~\cite{nabi2016machine}. The short-term dependency stream 1D CNN that has 128 kernels of size one and the he\_uniform function function as the kernel initializer. Then, the GRU number of units is set to 10, with glorot\_uniform function as the kernel initializer. The long-term dependency stream started with a similar 1D CNN followed by the LSTM that has ten units, and the kernel initialization function is glorot\_uniform and orthogonal function as the recurrent initializer. Then, the two streams concatenated, followed by two dense layers of 64 and 32, units with the rectified linear unit (ReLU)~\cite{teh2000rate,elsayed2018empirical} as the activation functions. For training the model, the batch size has been set to 32 and the number of epochs to 20. RMSProp has been used as the optimization function with learning rate $lr = 0.001$, $momentum = 0.0$, and $\epsilon = 1e-07$. The categorical cross-entropy is set as the loss function. Max pooling and dropouts have been applied in both the long-term and short-term dependency streams. Fig.~\ref{tess_train_validate} shows the proposed model training versus validation accuracy and loss. Table~\ref{results_tess_model} shows the proposed model's overall statistics over the TESS dataset. Table~\ref{class_level_tess} shows the statistics of the model over each of the seven emotions, where (OP) and (AUC) are the areas under the Receiver Operating Characteristic (ROC) curve. Table~\ref{tess_compare} compares the proposed model and the state-of-the-art speech emotion recognition models over the TESS dataset.

\section{Task II: Electrocardiogram Classification}
\subsection{Dataset Description}
In this task, we aim to empirically evaluate the TemporalAugmenter concept of the electrocardiogram (ECG) classification task as an example of a biological signal that carries information about heart functionality. In this experiment, we used the MIT-BIH dataset benchmark collected by the BIH Laboratory~\cite{moody2001impact}. The ECG signals were recorded from 25 women between 32 to 89 years old and 22 women aged between 23 to 89 years old. The dataset consists of 109,446 data samples. The dataset consists of five different categories of the ECG recorded signal, including normal heart beat (N), supraventricular premature beat (S), premature ventricular contraction (V), fusion of paced and normal beat (F), and unclassifiable beat (Q).

\subsection{Results and Analysis}
For the experiment, we divided into training, validation, and testing by the ratio 60\%, 20\%, and 20\%, respectively. The batch size is set to 128, and the number of epochs to 50. The first ratio of dropout layers at the long-term and short-term steams was set to 50\%, maintaining the rest at the 30\% ratio. Adam optimizer has been used with learning rate $lr = 0.001$, and $\epsilon = 1e-07$. The proposed model training versus validation accuracy and loss diagrams are shown in Figure~\ref{mih_train_val}. Table~\ref{results_mit_model} shows the proposed model's overall statistics over the MIT-BIH dataset. Table~\ref{class_level_mit} shows the statistics of the model over each of the five ECG categories. Table~\ref{mit_compare} compares the proposed model and the state-of-the-art classification models over the MIT-BIH dataset. 

\section{Task III: Radar Signal Quality Classification}

\subsection{Dataset Description}
In this experiment, we used the Ionosphere Depletion dataset benchmark collected by a Goose Bay system to evaluate the Ionosphere. The dataset consists of 351 data samples. The pulse numbers of the Good Bay system were 17. Each data sample in the dataset is described by two attributes per pulse number, corresponding to the complex values returned by the function resulting from the complex electromagnetic signal. Thus, the total number of attributes is 34. The data consists of two categories, bad signal and good signal, which are decoded to zero and one in the implementation. The dataset is unbalanced.

\subsection{Results and Analysis}
In this experiment, we split the data to train, validate, and test the dataset with ratios 60\%, 20\%, and 20\%, respectively. The Adam optimizer has been used with a  learning rate $lr = 0.001$, and $\epsilon = 1e-07$. The number of epochs is set to 100, and the batch size to 128. The proposed model overall statistics for the Radar Ionosphere Depletion dataset experiment are shown in Table~\ref{results_radar_model}. The proposed model statistics over the two categories of the Radar Ionosphere Depletion dataset are shown in Table~\ref{class_level_radar}. Table~\ref{radar_compare} compares the proposed model and the state-of-the-art Radar Ionosphere Depletion dataset classification models. Our proposed model exceeds the state-of-the-art models' accuracy performance.

\section{Conclusion and Social Impact}
Manipulating temporal data requires robust methodologies that can capture both temporal and point-to-time information. Several applications are based on classifying temporal data, such as biosignals classification for diagnostics, speech emotion recognition, stock market prediction, energy consumption, time series anomaly detection, and radar signal classification. This paper proposed a simple ensemble based approach, TemporalAugmenter, based on integrating long-term and short-term dependency learning streams to capture precise temporal dependencies in temporal data. In addition, we found empirically that one convolutional layer prior to a recurrent architecture can help enhance the model performance. Furthermore, the proposed approach can be used directly without extracting additional features, preprocessing, or converting the signal to spectrograms, which reduces the power and energy required for model implementation, contributing towards green AI and reducing CO$\mathrm{_{2}}$ footprint. Thus, models that are based on the proposed TemporalAugmenter approach can be implemented within different temporal data based application domains.

%Human speech is a complex biosignal with lexical, emotional, and biometric context. Recognizing different speech emotions can help understand and identify the reasons behind human decisions and behavior. Integrating human emotions in different human-computer interaction (HCI) research domains can help to increase user satisfaction and improve the user experience. However, embedding speech emotion requires additional computations to the existing systems. This paper proposed a simple ensemble-based approach, TemporalAugmenter, based on integrating long-term and short-term dependencies learning streams to capture precise temporal dependencies in speech for emotion recognition from emotion detection. In addition, we empirically found that the pooling layers do not provide improvements with unbalanced data categories in speech signal data. Moreover, we found empirically that one convolutional layer with kernels of size one prior to a recurrent architecture can help enhance the model performance. Furthermore, the proposed approach can be used directly without extracting additional features, preprocessing, or converting spectrograms, which reduces the power and energy required for model implementation, contributing towards green AI and reducing CO$\mathrm{_{2}}$ footprint. Thus, models that are based on the proposed TemporalAugmenter approach can be implemented within different HCI application domains. 

\bibliographystyle{flairs}
\bibliography{references}

\begin{thebibliography}{}

\bibitem[\protect\citeauthoryear{Acharya \bgroup et al\mbox.\egroup
  }{2017}]{acharya2017deep}
Acharya, U.~R.; Oh, S.~L.; Hagiwara, Y.; Tan, J.~H.; Adam, M.; Gertych, A.; and
  San~Tan, R.
\newblock 2017.
\newblock A deep convolutional neural network model to classify heartbeats.
\newblock {\em Computers in biology and medicine} 89:389--396.

\bibitem[\protect\citeauthoryear{Adewopo and Elsayed}{2023}]{adewopo2023smart}
Adewopo, V., and Elsayed, N.
\newblock 2023.
\newblock Smart city transportation: Deep learning ensemble approach for
  traffic accident detection.
\newblock {\em arXiv preprint arXiv:2310.10038}.

\bibitem[\protect\citeauthoryear{Al-Shabandar \bgroup et al\mbox.\egroup
  }{2021}]{al2021deep}
Al-Shabandar, R.; Jaddoa, A.; Liatsis, P.; and Hussain, A.~J.
\newblock 2021.
\newblock A deep gated recurrent neural network for petroleum production
  forecasting.
\newblock {\em Machine Learning with Applications} 3:100013.

\bibitem[\protect\citeauthoryear{Alotaibi and
  Ilyas}{2023}]{alotaibi2023ensemble}
Alotaibi, Y., and Ilyas, M.
\newblock 2023.
\newblock Ensemble-learning framework for intrusion detection to enhance
  internet of things’ devices security.
\newblock {\em Sensors} 23(12):5568.

\bibitem[\protect\citeauthoryear{Altman and
  Krzywinski}{2017}]{altman2017ensemble}
Altman, N., and Krzywinski, M.
\newblock 2017.
\newblock Ensemble methods: bagging and random forests.
\newblock {\em Nature Methods} 14(10):933--935.

\bibitem[\protect\citeauthoryear{Arpit \bgroup et al\mbox.\egroup
  }{2022}]{arpit2022ensemble}
Arpit, D.; Wang, H.; Zhou, Y.; and Xiong, C.
\newblock 2022.
\newblock Ensemble of averages: Improving model selection and boosting
  performance in domain generalization.
\newblock {\em Advances in Neural Information Processing Systems}
  35:8265--8277.

\bibitem[\protect\citeauthoryear{Asl, Setarehdan, and
  Mohebbi}{2008}]{asl2008support}
Asl, B.~M.; Setarehdan, S.~K.; and Mohebbi, M.
\newblock 2008.
\newblock Support vector machine-based arrhythmia classification using reduced
  features of heart rate variability signal.
\newblock {\em Artificial intelligence in medicine} 44(1):51--64.

\bibitem[\protect\citeauthoryear{Basheer \bgroup et al\mbox.\egroup
  }{2024}]{basheer2024autonomous}
Basheer, M. Y.~I.; Ali, A.~M.; Hamid, N. H.~A.; Ariffin, M. A.~M.; Osman, R.;
  Nordin, S.; and Gu, X.
\newblock 2024.
\newblock Autonomous anomaly detection for streaming data.
\newblock {\em Knowledge-Based Systems} 284:111235.

\bibitem[\protect\citeauthoryear{Bhargava and
  Polzehl}{2013}]{bhargava2013improving}
Bhargava, M., and Polzehl, T.
\newblock 2013.
\newblock Improving automatic emotion recognition from speech using rhythm and
  temporal feature.
\newblock {\em arXiv preprint arXiv:1303.1761}.

\bibitem[\protect\citeauthoryear{Brownlee}{2021}]{brownlee2021ensemble}
Brownlee, J.
\newblock 2021.
\newblock {\em Ensemble learning algorithms with Python: Make better
  predictions with bagging, boosting, and stacking}.
\newblock Machine Learning Mastery.

\bibitem[\protect\citeauthoryear{Chmielewski and
  Grzymala-Busse}{1996}]{chmielewski1996global}
Chmielewski, M.~R., and Grzymala-Busse, J.~W.
\newblock 1996.
\newblock Global discretization of continuous attributes as preprocessing for
  machine learning.
\newblock {\em International journal of approximate reasoning} 15(4):319--331.

\bibitem[\protect\citeauthoryear{Chung \bgroup et al\mbox.\egroup
  }{2014}]{chung2014empirical}
Chung, J.; Gulcehre, C.; Cho, K.; and Bengio, Y.
\newblock 2014.
\newblock Empirical evaluation of gated recurrent neural networks on sequence
  modeling.
\newblock {\em arXiv preprint arXiv:1412.3555}.

\bibitem[\protect\citeauthoryear{Dey and Salem}{2017}]{dey2017gate}
Dey, R., and Salem, F.~M.
\newblock 2017.
\newblock Gate-variants of gated recurrent unit (gru) neural networks.
\newblock In {\em 2017 IEEE 60th international midwest symposium on circuits
  and systems (MWSCAS)},  1597--1600.
\newblock IEEE.

\bibitem[\protect\citeauthoryear{Dietterich}{2002}]{dietterich2002machine}
Dietterich, T.~G.
\newblock 2002.
\newblock Machine learning for sequential data: A review.
\newblock In {\em Structural, Syntactic, and Statistical Pattern Recognition:
  Joint IAPR International Workshops SSPR 2002 and SPR 2002 Windsor, Ontario,
  Canada, August 6--9, 2002 Proceedings},  15--30.
\newblock Springer.

\bibitem[\protect\citeauthoryear{Drucker \bgroup et al\mbox.\egroup
  }{1994}]{drucker1994boosting}
Drucker, H.; Cortes, C.; Jackel, L.~D.; LeCun, Y.; and Vapnik, V.
\newblock 1994.
\newblock Boosting and other ensemble methods.
\newblock {\em Neural computation} 6(6):1289--1301.

\bibitem[\protect\citeauthoryear{Dupuis and
  Pichora-Fuller}{2010}]{dupuis2010toronto}
Dupuis, K., and Pichora-Fuller, M.~K.
\newblock 2010.
\newblock Toronto emotional speech set {(TESS)}-younger talker\_happy.

\bibitem[\protect\citeauthoryear{Elsayed and
  Zaghloul}{2020}]{elsayed2020simple}
Elsayed, N., and Zaghloul, Z.~S.
\newblock 2020.
\newblock A simple extreme learning machine model for detecting heart
  arrhythmia in the electrocardiography signal.
\newblock In {\em 2020 IEEE 63rd International Midwest Symposium on Circuits
  and Systems (MWSCAS)},  513--516.
\newblock IEEE.

\bibitem[\protect\citeauthoryear{Elsayed \bgroup et al\mbox.\egroup
  }{2022}]{elsayed2022speech}
Elsayed, N.; ElSayed, Z.; Asadizanjani, N.; Ozer, M.; Abdelgawad, A.; and
  Bayoumi, M.
\newblock 2022.
\newblock Speech emotion recognition using supervised deep recurrent system for
  mental health monitoring.
\newblock In {\em 2022 IEEE 8th World Forum on Internet of Things (WF-IoT)},
  1--6.
\newblock IEEE.

\bibitem[\protect\citeauthoryear{Elsayed, ElSayed, and
  Bayoumi}{2023}]{elsayed2023iot}
Elsayed, N.; ElSayed, Z.; and Bayoumi, M.
\newblock 2023.
\newblock Iot botnet detection using an economic deep learning model.
\newblock {\em arXiv preprint arXiv:2302.02013}.

\bibitem[\protect\citeauthoryear{Elsayed, ElSayed, and
  Maida}{2022}]{elsayed2022litelstm}
Elsayed, N.; ElSayed, Z.; and Maida, A.~S.
\newblock 2022.
\newblock Litelstm architecture for deep recurrent neural networks.
\newblock In {\em 2022 IEEE International Symposium on Circuits and Systems
  (ISCAS)},  1304--1308.
\newblock IEEE.

\bibitem[\protect\citeauthoryear{Elsayed, ElSayed, and
  Maida}{2023}]{elsayed2023litelstm}
Elsayed, N.; ElSayed, Z.; and Maida, A.~S.
\newblock 2023.
\newblock Litelstm architecture based on weights sharing for recurrent neural
  networks.
\newblock {\em arXiv preprint arXiv:2301.04794}.

\bibitem[\protect\citeauthoryear{Elsayed, Maida, and
  Bayoumi}{2018a}]{elsayed2018deep}
Elsayed, N.; Maida, A.~S.; and Bayoumi, M.
\newblock 2018a.
\newblock Deep gated recurrent and convolutional network hybrid model for
  univariate time series classification.
\newblock {\em arXiv preprint arXiv:1812.07683}.

\bibitem[\protect\citeauthoryear{Elsayed, Maida, and
  Bayoumi}{2018b}]{elsayed2018empirical}
Elsayed, N.; Maida, A.~S.; and Bayoumi, M.
\newblock 2018b.
\newblock Empirical activation function effects on unsupervised convolutional
  lstm learning.
\newblock In {\em 2018 IEEE 30th International Conference on Tools with
  Artificial Intelligence (ICTAI)},  336--343.
\newblock IEEE.

\bibitem[\protect\citeauthoryear{Elsayed, Maida, and
  Bayoumi}{2019}]{elsayed2019gated}
Elsayed, N.; Maida, A.~S.; and Bayoumi, M.
\newblock 2019.
\newblock Gated recurrent neural networks empirical utilization for time series
  classification.
\newblock In {\em 2019 International Conference on Internet of Things (iThings)
  and IEEE Green Computing and Communications (GreenCom) and IEEE Cyber,
  Physical and Social Computing (CPSCom) and IEEE Smart Data (SmartData)},
  1207--1210.
\newblock IEEE.

\bibitem[\protect\citeauthoryear{Elsayed, Zaghloul, and
  Li}{2021}]{elsayed2021arrhythmia}
Elsayed, N.; Zaghloul, Z.~S.; and Li, C.
\newblock 2021.
\newblock Arrhythmia supraventricular premature beat detection in
  electrocardiography signal using deep gated recurrent model.
\newblock In {\em SoutheastCon 2021},  1--7.
\newblock IEEE.

\bibitem[\protect\citeauthoryear{Ferreira and
  Figueiredo}{2012}]{ferreira2012boosting}
Ferreira, A.~J., and Figueiredo, M.~A.
\newblock 2012.
\newblock Boosting algorithms: A review of methods, theory, and applications.
\newblock {\em Ensemble machine learning: Methods and applications}  35--85.

\bibitem[\protect\citeauthoryear{Galar \bgroup et al\mbox.\egroup
  }{2011}]{galar2011review}
Galar, M.; Fernandez, A.; Barrenechea, E.; Bustince, H.; and Herrera, F.
\newblock 2011.
\newblock A review on ensembles for the class imbalance problem: bagging-,
  boosting-, and hybrid-based approaches.
\newblock {\em IEEE Transactions on Systems, Man, and Cybernetics, Part C
  (Applications and Reviews)} 42(4):463--484.

\bibitem[\protect\citeauthoryear{Ganaie \bgroup et al\mbox.\egroup
  }{2022}]{ganaie2022ensemble}
Ganaie, M.~A.; Hu, M.; Malik, A.; Tanveer, M.; and Suganthan, P.
\newblock 2022.
\newblock Ensemble deep learning: A review.
\newblock {\em Engineering Applications of Artificial Intelligence} 115:105151.

\bibitem[\protect\citeauthoryear{Gao, Zheng, and Guo}{2020}]{gao2020gated}
Gao, S.; Zheng, Y.; and Guo, X.
\newblock 2020.
\newblock Gated recurrent unit-based heart sound analysis for heart failure
  screening.
\newblock {\em Biomedical engineering online} 19:1--17.

\bibitem[\protect\citeauthoryear{Gers, Schraudolph, and
  Schmidhuber}{2002}]{gers2002learning}
Gers, F.~A.; Schraudolph, N.~N.; and Schmidhuber, J.
\newblock 2002.
\newblock Learning precise timing with lstm recurrent networks.
\newblock {\em Journal of machine learning research} 3(Aug):115--143.

\bibitem[\protect\citeauthoryear{Golmohammadi \bgroup et al\mbox.\egroup
  }{2017}]{golmohammadi2017gated}
Golmohammadi, M.; Ziyabari, S.; Shah, V.; Von~Weltin, E.; Campbell, C.; Obeid,
  I.; and Picone, J.
\newblock 2017.
\newblock Gated recurrent networks for seizure detection.
\newblock In {\em 2017 IEEE Signal Processing in Medicine and Biology Symposium
  (SPMB)},  1--5.
\newblock IEEE.

\bibitem[\protect\citeauthoryear{Greff \bgroup et al\mbox.\egroup
  }{2016}]{greff2016lstm}
Greff, K.; Srivastava, R.~K.; Koutn{\'\i}k, J.; Steunebrink, B.~R.; and
  Schmidhuber, J.
\newblock 2016.
\newblock Lstm: A search space odyssey.
\newblock {\em IEEE transactions on neural networks and learning systems}
  28(10):2222--2232.

\bibitem[\protect\citeauthoryear{Greff \bgroup et al\mbox.\egroup
  }{2017}]{greff2017lstm}
Greff, K.; Srivastava, R.~K.; Koutn{\'\i}k, J.; Steunebrink, B.~R.; and
  Schmidhuber, J.
\newblock 2017.
\newblock {LSTM}: A search space odyssey.
\newblock {\em IEEE Transactions on Neural Networks and Learning Systems}
  28(10):2222--2232.

\bibitem[\protect\citeauthoryear{Han, Chen, and Liu}{2021}]{han2021gan}
Han, X.; Chen, X.; and Liu, L.-P.
\newblock 2021.
\newblock Gan ensemble for anomaly detection.
\newblock In {\em Proceedings of the AAAI Conference on Artificial
  Intelligence}, volume~35,  4090--4097.

\bibitem[\protect\citeauthoryear{Hassan and Verma}{2007}]{hassan2007decisions}
Hassan, S.~Z., and Verma, B.
\newblock 2007.
\newblock Decisions fusion strategy: Towards hybrid cluster ensemble.
\newblock In {\em 2007 3rd International Conference on Intelligent Sensors,
  Sensor Networks and Information},  377--382.
\newblock IEEE.

\bibitem[\protect\citeauthoryear{Jakubik}{2018}]{jakubik2018evaluation}
Jakubik, J.
\newblock 2018.
\newblock Evaluation of gated recurrent neural networks in music classification
  tasks.
\newblock In {\em Information Systems Architecture and Technology: Proceedings
  of 38th International Conference on Information Systems Architecture and
  Technology--ISAT 2017: Part I},  27--37.
\newblock Springer.

\bibitem[\protect\citeauthoryear{Jia, Liang, and Liang}{2023}]{jia2023review}
Jia, J.; Liang, W.; and Liang, Y.
\newblock 2023.
\newblock A review of hybrid and ensemble in deep learning for natural language
  processing.
\newblock {\em arXiv preprint arXiv:2312.05589}.

\bibitem[\protect\citeauthoryear{Kachuee, Fazeli, and
  Sarrafzadeh}{2018}]{kachuee2018ecg}
Kachuee, M.; Fazeli, S.; and Sarrafzadeh, M.
\newblock 2018.
\newblock Ecg heartbeat classification: A deep transferable representation.
\newblock In {\em 2018 IEEE international conference on healthcare informatics
  (ICHI)},  443--444.
\newblock IEEE.

\bibitem[\protect\citeauthoryear{Karim, Majumdar, and
  Darabi}{2019}]{karim2019insights}
Karim, F.; Majumdar, S.; and Darabi, H.
\newblock 2019.
\newblock Insights into lstm fully convolutional networks for time series
  classification.
\newblock {\em IEEE Access} 7:67718--67725.

\bibitem[\protect\citeauthoryear{Kourentzes, Barrow, and
  Crone}{2014}]{kourentzes2014neural}
Kourentzes, N.; Barrow, D.~K.; and Crone, S.~F.
\newblock 2014.
\newblock Neural network ensemble operators for time series forecasting.
\newblock {\em Expert Systems with Applications} 41(9):4235--4244.

\bibitem[\protect\citeauthoryear{Krishnan, Joseph~Raj, and
  Rajangam}{2021}]{krishnan2021emotion}
Krishnan, P.~T.; Joseph~Raj, A.~N.; and Rajangam, V.
\newblock 2021.
\newblock Emotion classification from speech signal based on empirical mode
  decomposition and non-linear features: Speech emotion recognition.
\newblock {\em Complex \& Intelligent Systems} 7:1919--1934.

\bibitem[\protect\citeauthoryear{Liu \bgroup et al\mbox.\egroup
  }{2019}]{liu2019ensembles}
Liu, C.; Ta, C.~N.; Rogers, J.~R.; Li, Z.; Lee, J.; Butler, A.~M.; Shang, N.;
  Kury, F. S.~P.; Wang, L.; Shen, F.; et~al.
\newblock 2019.
\newblock Ensembles of natural language processing systems for portable
  phenotyping solutions.
\newblock {\em Journal of biomedical informatics} 100:103318.

\bibitem[\protect\citeauthoryear{Liu \bgroup et al\mbox.\egroup
  }{2020}]{liu2020novel}
Liu, N.; Li, X.; Qi, E.; Xu, M.; Li, L.; and Gao, B.
\newblock 2020.
\newblock A novel ensemble learning paradigm for medical diagnosis with
  imbalanced data.
\newblock {\em IEEE Access} 8:171263--171280.

\bibitem[\protect\citeauthoryear{Lotfidereshgi and
  Gournay}{2017}]{lotfidereshgi2017biologically}
Lotfidereshgi, R., and Gournay, P.
\newblock 2017.
\newblock Biologically inspired speech emotion recognition.
\newblock In {\em 2017 IEEE international conference on acoustics, speech and
  signal processing (ICASSP)},  5135--5139.
\newblock IEEE.

\bibitem[\protect\citeauthoryear{Martis \bgroup et al\mbox.\egroup
  }{2013a}]{martis2013application}
Martis, R.~J.; Acharya, U.~R.; Lim, C.~M.; Mandana, K.; Ray, A.~K.; and
  Chakraborty, C.
\newblock 2013a.
\newblock Application of higher order cumulant features for cardiac health
  diagnosis using ecg signals.
\newblock {\em International journal of neural systems} 23(04):1350014.

\bibitem[\protect\citeauthoryear{Martis \bgroup et al\mbox.\egroup
  }{2013b}]{martis2013cardiac}
Martis, R.~J.; Acharya, U.~R.; Mandana, K.; Ray, A.~K.; and Chakraborty, C.
\newblock 2013b.
\newblock Cardiac decision making using higher order spectra.
\newblock {\em Biomedical Signal Processing and Control} 8(2):193--203.

\bibitem[\protect\citeauthoryear{Moody and Mark}{2001}]{moody2001impact}
Moody, G.~B., and Mark, R.~G.
\newblock 2001.
\newblock The impact of the mit-bih arrhythmia database.
\newblock {\em IEEE engineering in medicine and biology magazine} 20(3):45--50.

\bibitem[\protect\citeauthoryear{Mosavi \bgroup et al\mbox.\egroup
  }{2021}]{mosavi2021ensemble}
Mosavi, A.; Sajedi~Hosseini, F.; Choubin, B.; Goodarzi, M.; Dineva, A.~A.; and
  Rafiei~Sardooi, E.
\newblock 2021.
\newblock Ensemble boosting and bagging based machine learning models for
  groundwater potential prediction.
\newblock {\em Water Resources Management} 35:23--37.

\bibitem[\protect\citeauthoryear{Nabi and Nabi}{2016}]{nabi2016machine}
Nabi, Z., and Nabi, Z.
\newblock 2016.
\newblock Machine learning at scale.
\newblock {\em Pro Spark Streaming: The Zen of Real-Time Analytics Using Apache
  Spark}  177--198.

\bibitem[\protect\citeauthoryear{Nanni \bgroup et al\mbox.\egroup
  }{2014}]{nanni2014ensemble}
Nanni, L.; Lumini, A.; Dominio, F.; Donadeo, M.; and Zanuttigh, P.
\newblock 2014.
\newblock Ensemble to improve gesture recognition.
\newblock {\em International Journal of Automated Identification Technology}.

\bibitem[\protect\citeauthoryear{Ortega, Caba{\~n}as, and
  Masegosa}{2022}]{ortega2022diversity}
Ortega, L.~A.; Caba{\~n}as, R.; and Masegosa, A.
\newblock 2022.
\newblock Diversity and generalization in neural network ensembles.
\newblock In {\em International Conference on Artificial Intelligence and
  Statistics},  11720--11743.
\newblock PMLR.

\bibitem[\protect\citeauthoryear{Osowski and Linh}{2001}]{osowski2001ecg}
Osowski, S., and Linh, T.~H.
\newblock 2001.
\newblock Ecg beat recognition using fuzzy hybrid neural network.
\newblock {\em IEEE Transactions on Biomedical Engineering} 48(11):1265--1271.

\bibitem[\protect\citeauthoryear{Pan \bgroup et al\mbox.\egroup
  }{2020}]{pan2020water}
Pan, M.; Zhou, H.; Cao, J.; Liu, Y.; Hao, J.; Li, S.; and Chen, C.-H.
\newblock 2020.
\newblock Water level prediction model based on gru and cnn.
\newblock {\em Ieee Access} 8:60090--60100.

\bibitem[\protect\citeauthoryear{Parry \bgroup et al\mbox.\egroup
  }{2019}]{parry19_interspeech}
Parry, J.; Palaz, D.; Clarke, G.; Lecomte, P.; Mead, R.; Berger, M.; and Hofer,
  G.
\newblock 2019.
\newblock {Analysis of Deep Learning Architectures for Cross-Corpus Speech
  Emotion Recognition}.
\newblock In {\em Proc. Interspeech 2019},  1656--1660.

\bibitem[\protect\citeauthoryear{Ponti~Jr}{2011}]{ponti2011combining}
Ponti~Jr, M.~P.
\newblock 2011.
\newblock Combining classifiers: from the creation of ensembles to the decision
  fusion.
\newblock In {\em 2011 24th SIBGRAPI Conference on Graphics, Patterns, and
  Images Tutorials},  1--10.
\newblock IEEE.

\bibitem[\protect\citeauthoryear{Priyadharshini \bgroup et al\mbox.\egroup
  }{2023}]{priyadharshini2023hybrid}
Priyadharshini, M.; Banu, A.~F.; Sharma, B.; Chowdhury, S.; Rabie, K.; and
  Shongwe, T.
\newblock 2023.
\newblock Hybrid multi-label classification model for medical applications
  based on adaptive synthetic data and ensemble learning.
\newblock {\em Sensors} 23(15):6836.

\bibitem[\protect\citeauthoryear{Sagi and Rokach}{2018}]{sagi2018ensemble}
Sagi, O., and Rokach, L.
\newblock 2018.
\newblock Ensemble learning: A survey.
\newblock {\em Wiley Interdisciplinary Reviews: Data Mining and Knowledge
  Discovery} 8(4):e1249.

\bibitem[\protect\citeauthoryear{Sajjad \bgroup et al\mbox.\egroup
  }{2020}]{sajjad2020novel}
Sajjad, M.; Khan, Z.~A.; Ullah, A.; Hussain, T.; Ullah, W.; Lee, M.~Y.; and
  Baik, S.~W.
\newblock 2020.
\newblock A novel cnn-gru-based hybrid approach for short-term residential load
  forecasting.
\newblock {\em Ieee Access} 8:143759--143768.

\bibitem[\protect\citeauthoryear{Sangamnerkar \bgroup et al\mbox.\egroup
  }{2020}]{sangamnerkar2020ensemble}
Sangamnerkar, S.; Srinivasan, R.; Christhuraj, M.; and Sukumaran, R.
\newblock 2020.
\newblock An ensemble technique to detect fabricated news article using machine
  learning and natural language processing techniques.
\newblock In {\em 2020 International Conference for Emerging Technology
  (INCET)},  1--7.
\newblock IEEE.

\bibitem[\protect\citeauthoryear{Seijo-Pardo \bgroup et al\mbox.\egroup
  }{2017}]{seijo2017ensemble}
Seijo-Pardo, B.; Porto-D{\'\i}az, I.; Bol{\'o}n-Canedo, V.; and
  Alonso-Betanzos, A.
\newblock 2017.
\newblock Ensemble feature selection: homogeneous and heterogeneous approaches.
\newblock {\em Knowledge-Based Systems} 118:124--139.

\bibitem[\protect\citeauthoryear{Shen \bgroup et al\mbox.\egroup
  }{2018}]{shen2018deep}
Shen, G.; Tan, Q.; Zhang, H.; Zeng, P.; and Xu, J.
\newblock 2018.
\newblock Deep learning with gated recurrent unit networks for financial
  sequence predictions.
\newblock {\em Procedia computer science} 131:895--903.

\bibitem[\protect\citeauthoryear{Sigillito \bgroup et al\mbox.\egroup
  }{1989}]{sigillito1989classification}
Sigillito, V.~G.; Wing, S.~P.; Hutton, L.~V.; and Baker, K.~B.
\newblock 1989.
\newblock Classification of radar returns from the ionosphere using neural
  networks.
\newblock {\em Johns Hopkins APL Technical Digest} 10(3):262--266.

\bibitem[\protect\citeauthoryear{Sundarprasad}{2018}]{sundarprasad2018speech}
Sundarprasad, N.
\newblock 2018.
\newblock Speech emotion detection using machine learning techniques.

\bibitem[\protect\citeauthoryear{Teh and Hinton}{2000}]{teh2000rate}
Teh, Y.~W., and Hinton, G.~E.
\newblock 2000.
\newblock Rate-coded restricted boltzmann machines for face recognition.
\newblock {\em Advances in neural information processing systems} 13.

\bibitem[\protect\citeauthoryear{Tsogbaatar \bgroup et al\mbox.\egroup
  }{2021}]{tsogbaatar2021iot}
Tsogbaatar, E.; Bhuyan, M.~H.; Taenaka, Y.; Fall, D.; Gonchigsumlaa, K.;
  Elmroth, E.; and Kadobayashi, Y.
\newblock 2021.
\newblock Del-iot: A deep ensemble learning approach to uncover anomalies in
  iot.
\newblock {\em Internet of Things} 14:100391.

\bibitem[\protect\citeauthoryear{Vanerio and Casas}{2017}]{vanerio2017ensemble}
Vanerio, J., and Casas, P.
\newblock 2017.
\newblock Ensemble-learning approaches for network security and anomaly
  detection.
\newblock In {\em Proceedings of the Workshop on Big Data Analytics and Machine
  Learning for Data Communication Networks},  1--6.

\bibitem[\protect\citeauthoryear{Venkataramanan and
  Rajamohan}{2019}]{venkataramanan2019emotion}
Venkataramanan, K., and Rajamohan, H.~R.
\newblock 2019.
\newblock Emotion recognition from speech.
\newblock {\em arXiv preprint arXiv:1912.10458}.

\bibitem[\protect\citeauthoryear{Wang \bgroup et al\mbox.\egroup
  }{2021}]{wang2021speech}
Wang, Y.; Han, J.; Zhang, T.; and Qing, D.
\newblock 2021.
\newblock Speech enhancement from fused features based on deep neural network
  and gated recurrent unit network.
\newblock {\em EURASIP Journal on Advances in Signal Processing} 2021:1--19.

\bibitem[\protect\citeauthoryear{West \bgroup et al\mbox.\egroup
  }{2005}]{west2005ensemble}
West, D.; Mangiameli, P.; Rampal, R.; and West, V.
\newblock 2005.
\newblock Ensemble strategies for a medical diagnostic decision support system:
  A breast cancer diagnosis application.
\newblock {\em European Journal of Operational Research} 162(2):532--551.

\bibitem[\protect\citeauthoryear{Xu \bgroup et al\mbox.\egroup
  }{2018}]{xu2018ensemble}
Xu, Y.; Cheng, J.; Wang, L.; Xia, H.; Liu, F.; and Tao, D.
\newblock 2018.
\newblock Ensemble one-dimensional convolution neural networks for
  skeleton-based action recognition.
\newblock {\em IEEE Signal Processing Letters} 25(7):1044--1048.

\bibitem[\protect\citeauthoryear{Yi{\u{g}}it \bgroup et al\mbox.\egroup
  }{2021}]{yiugit2021automatic}
Yi{\u{g}}it, E.; {\"O}zkaya, U.; {\"O}zt{\"u}rk, {\c{S}}.; Singh, D.; and
  Gritli, H.
\newblock 2021.
\newblock Automatic detection of power quality disturbance using convolutional
  neural network structure with gated recurrent unit.
\newblock {\em Mobile Information Systems} 2021:1--11.

\bibitem[\protect\citeauthoryear{Yu \bgroup et al\mbox.\egroup
  }{2020}]{yu2020deep}
Yu, X.; Zhang, Z.; Wu, L.; Pang, W.; Chen, H.; Yu, Z.; and Li, B.
\newblock 2020.
\newblock Deep ensemble learning for human action recognition in still images.
\newblock {\em Complexity} 2020:1--23.

\bibitem[\protect\citeauthoryear{Zhang, Cheng, and
  Hsieh}{2019}]{zhang2019enhancing}
Zhang, H.; Cheng, M.; and Hsieh, C.-J.
\newblock 2019.
\newblock Enhancing certifiable robustness via a deep model ensemble.
\newblock {\em arXiv preprint arXiv:1910.14655}.

\bibitem[\protect\citeauthoryear{Zhang, Zhao, and Lei}{2013}]{zhang2013speech}
Zhang, S.; Zhao, X.; and Lei, B.
\newblock 2013.
\newblock Speech emotion recognition using an enhanced kernel isomap for
  human-robot interaction.
\newblock {\em International Journal of Advanced Robotic Systems} 10(2):114.

\bibitem[\protect\citeauthoryear{Zhao, Mao, and Chen}{2019}]{zhao2019speech}
Zhao, J.; Mao, X.; and Chen, L.
\newblock 2019.
\newblock Speech emotion recognition using deep 1d \& 2d cnn lstm networks.
\newblock {\em Biomedical signal processing and control} 47:312--323.

\bibitem[\protect\citeauthoryear{Zhao, Mehrotra, and
  Mohan}{2015}]{zhao2015ensemble}
Zhao, Z.; Mehrotra, K.~G.; and Mohan, C.~K.
\newblock 2015.
\newblock Ensemble algorithms for unsupervised anomaly detection.
\newblock In {\em Current Approaches in Applied Artificial Intelligence: 28th
  International Conference on Industrial, Engineering and Other Applications of
  Applied Intelligent Systems, IEA/AIE 2015, Seoul, South Korea, June 10-12,
  2015, Proceedings 28},  514--525.
\newblock Springer.

\bibitem[\protect\citeauthoryear{Zhou \bgroup et al\mbox.\egroup
  }{2016}]{zhou2016minimal}
Zhou, G.-B.; Wu, J.; Zhang, C.-L.; and Zhou, Z.-H.
\newblock 2016.
\newblock Minimal gated unit for recurrent neural networks.
\newblock {\em International Journal of Automation and Computing}
  13(3):226--234.

\bibitem[\protect\citeauthoryear{Zhou, Jin, and Dong}{2017}]{zhou2017premature}
Zhou, F.-y.; Jin, L.-p.; and Dong, J.
\newblock 2017.
\newblock Premature ventricular contraction detection combining deep neural
  networks and rules inference.
\newblock {\em Artificial intelligence in medicine} 79:42--51.

\end{thebibliography}

\end{document}